%% file: main.tex
\documentclass[conference]{IEEEtran}
\usepackage{times}
\usepackage[final]{changes}
\usepackage{enumitem}

\usepackage{booktabs}
\usepackage[normalem]{ulem}
\useunder{\uline}{\ul}{}
\usepackage[numbers]{natbib}
\usepackage{multicol}
\usepackage{booktabs}
\usepackage[bookmarks=true]{hyperref}
\usepackage{graphicx}
\usepackage{amsmath}
\usepackage{subcaption}
\usepackage{xcolor}
\usepackage[font=small,labelfont=bf,tableposition=top]{caption}
\usepackage{multirow}
\usepackage{bm}
\usepackage{enumitem}
\usepackage{xspace}
\usepackage{pdfx}

\newcommand{\shortname}[0]{Vid2Robot\xspace}

\begin{document}
\title{\shortname: End-to-end Video-conditioned Policy Learning with Cross-Attention Transformers}


\author{\small
Vidhi Jain$^\text{1,2}$ \quad  Maria Attarian$^\text{1,3}$ \quad  Nikhil J Joshi$^\text{1}$ \quad  Ayzaan Wahid$^\text{1}$ \quad  Danny Driess$^\text{1}$ \quad  
Quan Vuong$^\text{1}$ \quad  Pannag R Sanketi$^\text{1}$ \quad  \\
Pierre Sermanet$^\text{1}$ \quad  Stefan Welker$^\text{1}$ \quad  Christine Chan$^\text{1}$ \quad  
Igor Gilitschenski$^\text{3}$ \quad  Yonatan Bisk$^\text{2}$ \quad  Debidatta Dwibedi$^\text{1}$
\vspace{-0.3em}
\\\\
\vspace{-0.2em}
$^\text{1}$Google DeepMind Robotics     \quad 
$^\text{2}$Carnegie Mellon University      \quad     
$^\text{3}$University of Toronto
}

\twocolumn[{%
\renewcommand\twocolumn[1][]{#1}%
\maketitle
\begin{center}
   \includegraphics[width=\linewidth]{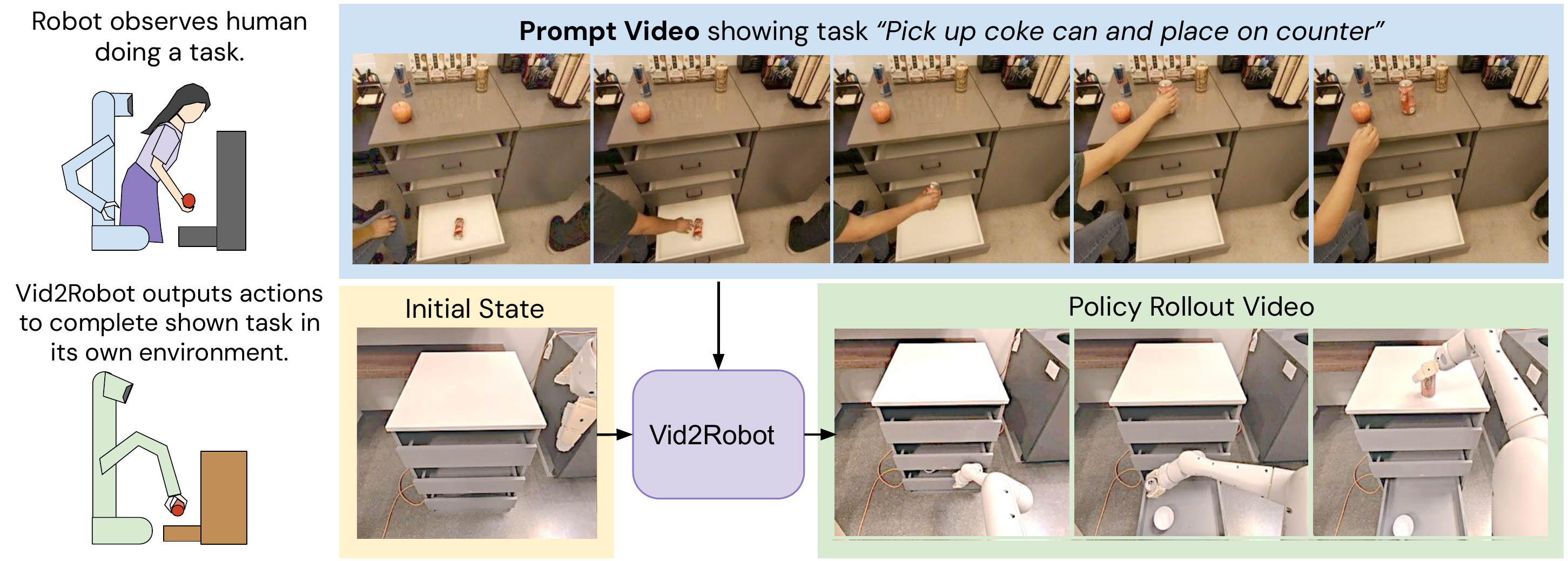}
   \captionof{figure}{\textbf{Overview.} Vid2Robot is a video-conditioned robot policy. Given a human demonstration (top), Vid2Robot recognizes the task semantics and performs the same task based on the robot's current visual observation (bottom left). A successful trajectory is presented on the bottom right.}
\label{fig:teaser}
\end{center}%
}]

\begin{abstract}
\input{sections/abstract}
\end{abstract}

\IEEEpeerreviewmaketitle

\input{sections/introduction}

\input{sections/approach}

\input{sections/experiments}
\input{sections/relatedwork}
\input{sections/limitation}

\input{sections/conclusion}
\input{sections/acknowledgements}

\appendix

\input{sections/appendix}

\bibliographystyle{plainnat}
\bibliography{references}

\end{document}

%% file: sections/abstract.tex
Large-scale multi-task robotic manipulation systems often rely on text to specify the task. In this work, we explore whether a robot can learn by observing humans. To do so, the robot must understand a person's intent and perform the inferred task despite differences in the embodiments and environments. 
We introduce \shortname{}, an end-to-end video-conditioned policy that takes
human videos demonstrating manipulation tasks as input and produces robot actions. Our model is trained with a large dataset of prompt video-robot trajectory pairs to learn unified representations of human and robot actions from videos.
\shortname{} uses cross-attention transformer layers 
between video features and the current robot state to produce the actions and perform the same task as shown in the video. We use auxiliary contrastive losses to align the prompt and robot video representations for better policies.
We evaluate \shortname{} on real-world robots and observe over 20\% improvement over BC-Z when using human prompt videos. Further, we also show cross-object motion transfer ability that enables video-conditioned policies to transfer a motion observed on one object in the prompt video to another object in the robot's own environment. 
Videos available at \textcolor{blue}{\href{https://vid2robot.github.io}{vid2robot.github.io}}.

%% file: sections/introduction.tex
\section{Introduction}
\label{sec:intro}

The path to creating versatile robots that assist in people's daily routines requires them to learn new skills on-the-go. These skills can vary from simple preferences for arranging dishes in the dishwasher in a specific household to completely different approaches to household cleaning. Humans can communicate in natural language for tasks that are already known. However, we revert to demonstrations when we want to learn a novel skill with nuance. For example, we might show how a particular microwave works or how to organize our cabinets. Robots need the same ability for generalization from demonstration, which comes so naturally to humans. 

Humans can infer the intentions of other humans based on third-person visual observations. Often, we use social reasoning and common sense to understand others' goals implicitly.
This ability is crucial for learning everyday tasks, such as kneading dough or knitting, where the intricacies are challenging to convey through still images or text~\cite{Bisk2020}. We often turn to How-To videos~\cite{miech19howto100m}) to learn how to perform such tasks. 
If robots could act based on videos, it would enable efficient task learning and improved interaction with humans.

This work focuses on visual imitation learning, where robots learn to perform tasks by watching video demonstrations. This setup offers several advantages. First, it allows robots to learn from agents with a different embodiment, enabling new skill acquisition without teleoperation. Second, 
it allows robots to infer tasks from expert demonstrations, even if the expert is not showing how to perform tasks in the same environment as the robot. Finally,
visual imitation learning is ideal for teaching tasks that are difficult or impossible to describe in words. 

Existing multi-task robot manipulation models (e.g. RT-1~\cite{Brohan2022-ta}, RT-2~\cite{Brohan2023-pe}, and RT-X~\cite{Open_X-Embodiment_Collaboration2023-ee}) use language conditioning to output a robot trajectory. Relying on text alone for task specification makes it difficult for robots to handle polysemy and tasks whose executions vary dramatically based on context.   
For example, `\textit{open} drawer', `\textit{open} cabinet', `\textit{open} container with lid' and `\textit{open} jar with screw cap' might share the same verb but require very different motor control for each interaction. Here, the agent should not generalize its policy, whereas it should generalize from one \textit{drawer} to others that vary in type, color, and shape.
For this reason, there are a broad range of tasks for which it is hard 
to design primitives for high-level planning approaches \cite{Liang2022-gl, arenas2023how}.

Another common approach has been to use a final goal image in goal-conditioned behavior cloning tasks \citep{Nair2018-nv, Krantz2023-qg}. However, \textit{how} to act is often ignored in such specifications. For example, `hold the flag' and `wave the flag' can have the same final goal image. We can resolve this ambiguity by using several sub-goal frames, \replaced{that is quite close to} {i.e.} conditioning robot policies with videos. 

\replaced{Current success of video conditioned policies in  ~\cite{Shah2023} assume that the provided video is from the same workspace with limited variability.}{Cases of good performance~\cite{Shah2023} with video conditioning require the provided video to be from the same workspace with limited variability.} 
Video-conditioned policies also lag in performance compared to language-conditioned policies work~\cite{Jang2022-mz}. 
Based on these and related work, we identify three main challenges for 
video-conditioned policies:
(1) Raw videos are high dimensional data that require \replaced{more computational power and memory}{more compute and memory to process}.
(2) 
While unlabeled video data are abundant on the internet, finding robot-specific video and motion data is hard. 
(3) 
People can perform the same task differently due to variations in objects, lighting conditions, and other background distractions.

Despite these challenges, video-conditioned policy learning is a core challenge robots need to master. To this end, we study how end-to-end models with video-conditioning can be used to specify tasks to the robot. 
Vid2Robot is an end-to-end system that enables rapid adaptation to tasks specified as video demonstrations. 
Unlike prior work that either learned representations from videos for only object and verb recognition \cite{Jang2022-mz} or learned motor control in simulation \cite{Xiao2022-by}, our work demonstrates the applicability of end-to-end learned video representations for real-world robotic control. 

We present the key contributions of our work as follows: 
(1) We present a transformer-based policy to encode video task specification, demonstrated by either robot or human agent embodiments (\S \ref{sec:approach}).
(2) We encourage alignment between the prompt and robot video representations using three contrastive auxiliary losses during training (\S \ref{subsec:training})
(3) Through actual robot experiments, we find our video-conditioned policy is better than baselines on human prompt videos. Furthermore, our policy is better at cross-object motion transfer (\S \ref{sec:exp}).

%% file: sections/approach.tex
\begin{figure*}[t]
\begin{center}
   \includegraphics[width=\linewidth]{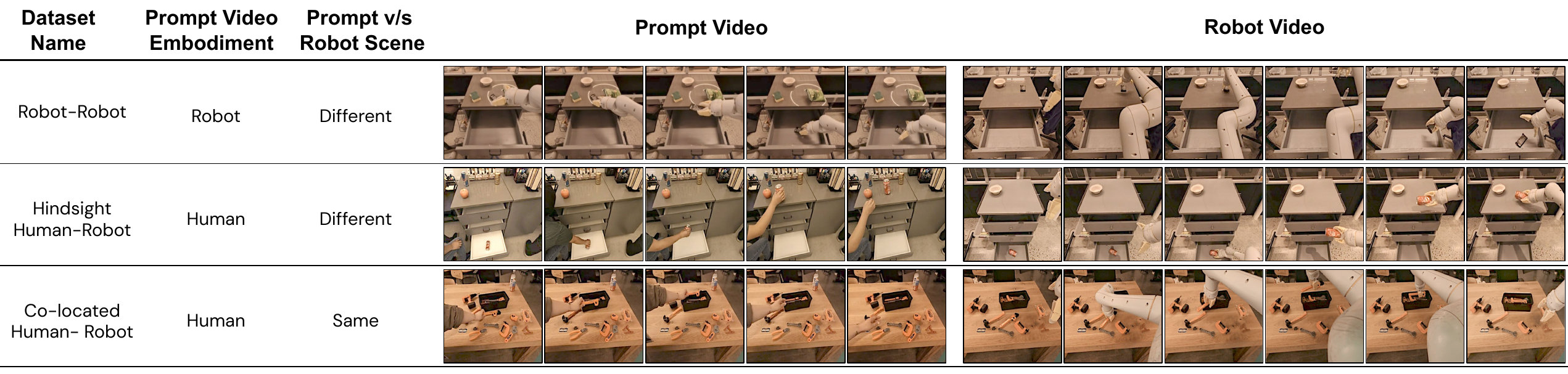}
\end{center}
   \caption{\textbf{Dataset creation.} (top row) Here we show a Robot-Robot video pair for \textit{placing the rxbar into top drawer}. We similarly pair existing robot-robot videos performing the same task.  (middle row) Here, we show Hindsight Human-Robot paired videos for \textit{picking a Coke can from the bottom drawer and placing it on the counter} task. We use the task instructions from robot trajectories, ask human participants to perform the task and record a demonstration video from the robot's perspective/view. (bottom row) Here, we show a Co-located Human-Robot pair of videos for \textit{placing the pipe wrench in the toolkit}. We record a human demonstration and a robot teleoperation in the same workspace. We use different workspaces to perform the same task instruction, thus collecting paired videos with visually diverse prompts and robot state observations. More details in Section~\ref{subsec:datasets}.}
\label{fig:dataset-creation}
\end{figure*}

\section{Approach}
\label{sec:approach}
\subsection{Preliminaries}
Our objective is to design a robotic system that takes in a \textit{prompt video} of a manipulation task and outputs actions that accomplish the task demonstrated in the video. This system must infer the underlying task from the prompt video (which might have a different setup or embodiment than the robot) and then manipulate the objects in its environment to achieve the inferred task.
Our model's inputs are a prompt video $V$ and the robot state $S_t = \{x_i\}_{i=t-k-1}^t$ where $x_i$ is the frame from the robot's camera stream at time $i$, $k$ is the maximum number of historical frames, and $t$ is the current timestep We train a policy $\pi(a_t | S_t, V)$ that infers the underlying task from $V$ and predicts task-relevant action $a_t$. 
We need a dataset of paired prompt videos and robot trajectories to train this model.
Below, we will discuss in detail how to create paired datasets.

\subsection{Datasets}
\label{subsec:datasets}

We need a dataset of pairs to train a video-conditioned robot policy: prompt videos and robot trajectories that perform the same task. In this work, we explore \replaced{prompt}{reference} videos where humans and robots perform the task. To create this dataset, we rely on three classes of data:

\textit{(1) Robot-Robot}: 
    We pair existing robot-robot videos of the same task. For this pairing, we consider two videos to \textit{match} if they perform the same task in different settings. We define ``\textit{task}'' based on natural language instructions for recording robot trajectories. These instructions typically consist of one or two verbs surrounded by nouns, such as `\textit{place} water bottle upright', `\textit{move} the coke can to the green chip bag' or `\textit{open} top drawer'. Note that we use language instructions only for pairing and as an input to the robot policy. The objective of this pairing is two-fold: first, to leverage the already labeled and collected dataset of robot trajectories, and second, to ensure robots can imitate the same task but in  different environment.

\textit{(2) Hindsight Human-Robot}: 
    Here, we use the task instructions from the robot trajectories dataset, ask one to five human participants to perform the task, and record a demonstration video from the robot's perspective/view. 
    The instructions are the same as before, but there is a significant embodiment and speed variability due to different humans performing the task with left or right hands and at a randomized robot camera angle. This provides us with a lot of paired data for training the policy with the available set of instructions in the robot dataset, without having to collect new robot trajectories. 

\textit{(3) Co-located Human-Robot}: 
    In this case, humans and robots perform the same task in the same workspace.
    We used this approach to collect human demonstrations and robot trajectories in diverse spaces such as a living space with sofas, a meeting room with whiteboards, a hardware workstations with toy tools, a kitchen with countertop, a refrigerator and a sink, a storage supplies area, and more.

We show examples of paired videos of the prompt and robot demo from each of the three datasets in Figure~\ref{fig:dataset-creation}. As can be seen, there is a considerable difference in the backgrounds and distractor objects in the Hindsight Human-Robot and Co-located Human-Robot datasets. A different complexity arises when comparing the first approach (Robot-Robot), where the actor is a robot with the same morphology, to the other two cases where the human is the actor in the prompt videos. 

Each source represents a different level of difficulty and time to collect. Pairing existing robot datasets requires no extra data collection but does not involve any human demonstrations. Our second data involves asking humans to mimic existing robot trajectories. Hindsight human videos made data collection easier as they do not need robot teleoperation data. However, this did not increase the diversity of tasks in the data set. Lastly, we collect data with humans and robots in the same environment. While collecting co-located paired data is a presumed gold standard, it is very time and labor-intensive compared to the previous two approaches. Thus, it forms a small fraction of our overall training set. After combining all the datasets,  we have $\sim$100k robot videos and $\sim$10k human videos covering the tasks introduced in RT-1~\cite{Brohan2022-ta} and RT-2~\cite{Brohan2023-pe}. The robot-robot dataset makes up more than 90\% of the entire dataset. This dataset is publicly available \cite{Open_X-Embodiment_Collaboration2023-ee}. We provide a Python Notebook in Supplementary Material for accessible visualization of the paired videos used in training.

\subsection{Model Architecture}
\label{subsec:architecture}

\begin{figure*}
\begin{center}
   \includegraphics[width=\linewidth]{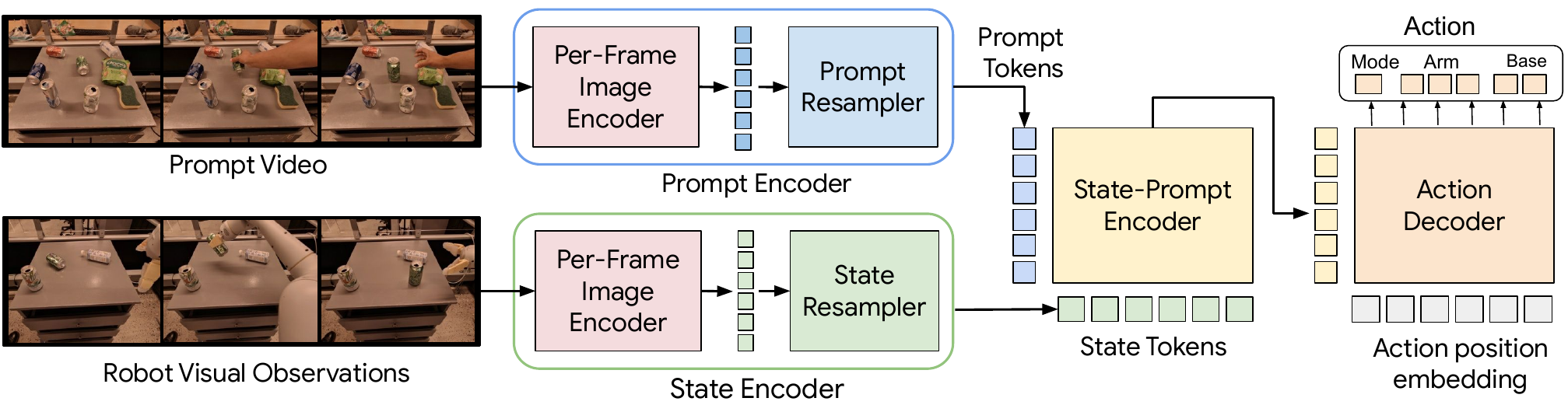}
\end{center}
   \caption{\textbf{Architecture.}
   Our model takes the prompt video and the robot's current observations as the input, encodes those into token embeddings for the prompt video and the robot's state, cross-attends to produce state-prompt encoding, and translates it into the expected robot action at the current timestep. More details in Section~\ref{subsec:architecture}).}
\label{fig:architecture}
\end{figure*}

Our policy takes the prompt video and the current robot state as inputs and outputs robot actions. It consists of four modules: (1) prompt video encoder, (2) robot state encoder, (3) state-prompt encoder, and (4) robot action decoder. The entire architecture is illustrated in Figure~\ref{fig:architecture}, and each of the modules is detailed below:

\textit{(1) Prompt Video Encoder} encodes the video demonstration provided as a reference to convey the desired task semantics. The prompt video encoder implicitly learns to infer what task to perform and how to do it. The prompt encoder consists of a per-frame Image encoder $\phi_p$ (ViT \cite{Dosovitskiy2020-xd})  followed by a Perceiver Resampler \cite{Alayrac2022-fm, Jaegle2021-rs} $\psi_p$. The output of the prompt encoder $\psi_p(\phi_p(V))=z_{prompt}$ is a set of $N$ tokens of d-dimension to condition the policy with the task-relevant attributes from the video. 

\textit{(2) Robot State Encoder }encodes the current state of the robot given the current frame and last $k$ frames as input. Note that this module also encodes information about the objects and environment visible to the robot. The architecture is similar to the prompt encoder: a per-frame Image encoder $\phi_s$ followed by a Perceiver Resampler $\psi_s$. Identical to the prompt encoder's outputs, the output of the state encoder is $\psi_s(\phi_s(S_t))=z_{state}$ that encodes the latent environment and robot state information from the history of recent observations. 
We use the same image encoder weights for both (1) and (2), that is, $\phi_p\!=\!\phi_s\!=\!\phi$.  The role of the image encoder $\phi$ is to 
capture spatial visual information in each frame.
The perceiver resampler enables temporal learning across frames and reduces the number of video tokens passed into the action decoder. 

\textit{(3) State-Prompt Encoder} takes the prompt video encoding $z_{prompt}$ and robot state encoding $z_{state}$ and outputs a task encoding relevant for action prediction, which we call prompt-aware state tokens $z_{state|prompt}$. Here, the state encoding acts as queries, and the prompt video encoding acts as keys and values. Intuitively, the state-prompt encoder enables the fusion of the state and prompt information. Suppose a prompt video shows picking up an apple in the basket, and the current state contains an apple, a banana, and an orange. The State-Prompt Encoder cross-attends and learns which object to attend to in the state based on the prompt video. Capturing interdependencies between prompt and state is crucial for the next step of action decoding. 

\textit{(4) Robot Action Decoder} predicts the action vector $a_t$ for the current state $S_t$ such that it completes the task shown in the prompt video $V_p$. 
The action decoder is a transformer decoder architecture that uses the fixed action position tokens \citep{Zhao2023-zd} as input queries and the prompt-aware state tokens $z_{state|prompt}$ for keys and values. The size of the action position embedding is $N \times d$ where $N$ is the number of action dimensions, and $d$ is the transformer embedding dimension.
More details on action vector in \S\ref{subsec:preprocessing}.%

The action position embeddings cross-attend to the prompt-aware state tokens to predict the target binned action values as output. Each output token of the action decoder corresponds to an action dimension for the mode, arm, and base. We project each token embedding to 256 dimensions, and a softmax layer is applied on the top to obtain the bin corresponding to the target action vector. Unlike prior work \cite{Brohan2022-ta, Brohan2023-pe} that use autoregressive action decoding that requires multiple forward passes during inference, we use action position embeddings for one forward pass prediction like in ACT \cite{Zhao2023-zd}. Instead of predicting one action for the next timestep, we follow the approach outlined in \cite{Jang2022-mz, Zhao2023-zd} and train the policy with a prediction horizon of four steps.  We always use the action bin with the highest probability, that is, \texttt{argmax} over predicted probabilities, to choose the action value for execution.

\textit{Cross-Attention Layers.} In the \shortname architecture, we use Cross-Attention Transformer layers extensively in the following modules: Prompt Resampler, State Resampler, State-Prompt Encoder, and Action Decoder. Compared to standard self-attention layers, which require more memory to process the same video, cross-attention layers help manage the high number of tokens and the resulting large attention matrices when processing prompt and robot state videos. For example, when using ViT-B/16, the total number of video tokens for a $16$ frame reference video and a $8$ frame robot state video at $224\times224$ resolution would be $8\times196 + 16\times 196 = 4704$. An entire self-attention operation would lead to an attention matrix with $4704^2 \sim 22\mathrm{M}$ entries. However, using two Perceiver Resamplers with 64 latent tokens, we train with attention matrices of the size $8\times196\times64 + 16\times196\times64 \sim .3 \mathrm{M}$. Thus, cross-attention layers in \shortname reduce attention computation and enable training with paired videos.

\subsection{Preprocessing}
\label{subsec:preprocessing}
To efficiently train videos of varying lengths, we randomly sample $N\!\!=\!\!16$ frames. We always include the first and last frames and sort them in increasing order of time. We sample a robot state $S_t$ during training by sampling a random timestep. We then select the preceding $k-1$ frames to create a robot state video comprising a total of $k\!\!=\!\!8$ frames before. If there are less than $k-1$ frames before the current time step, we repeat the first frame to create a fixed-size robot state video. 
We normalize the pixel values in each frame between 0 and 1 and resize each frame to $(224, 224)$. During training, we apply photometric distortions like cropping, brightness, contrast, hue, and saturation.

The action at that time consists of the three components:
\textit{Mode}: ($m$) whether to terminate episode, move only arm, move only base, or both.
\textit{arm}: gripper position (x, y, z), orientation (rotation along xy, yz, zx), and the degree of closedness (c).  
\textit{Base}: displacement (x, y) and rotation.
Overall, the action $a_t = [m, g_x, g_y, g_z, \theta_{xy}, \theta_{yz}, \theta_{zx}, c, b_x, b_y, b_{\theta}]$ is an 11-dimensional vector.
Each value has different ranges, which we first use to scale the values between 0 and 1. We further discretize the values into 256 bins each. 
In this study, we train and evaluate scenarios where the base remains stationary.

\subsection{Training}
\label{subsec:training}
\begin{figure*}[t]
\begin{center}
   \includegraphics[width=\linewidth]{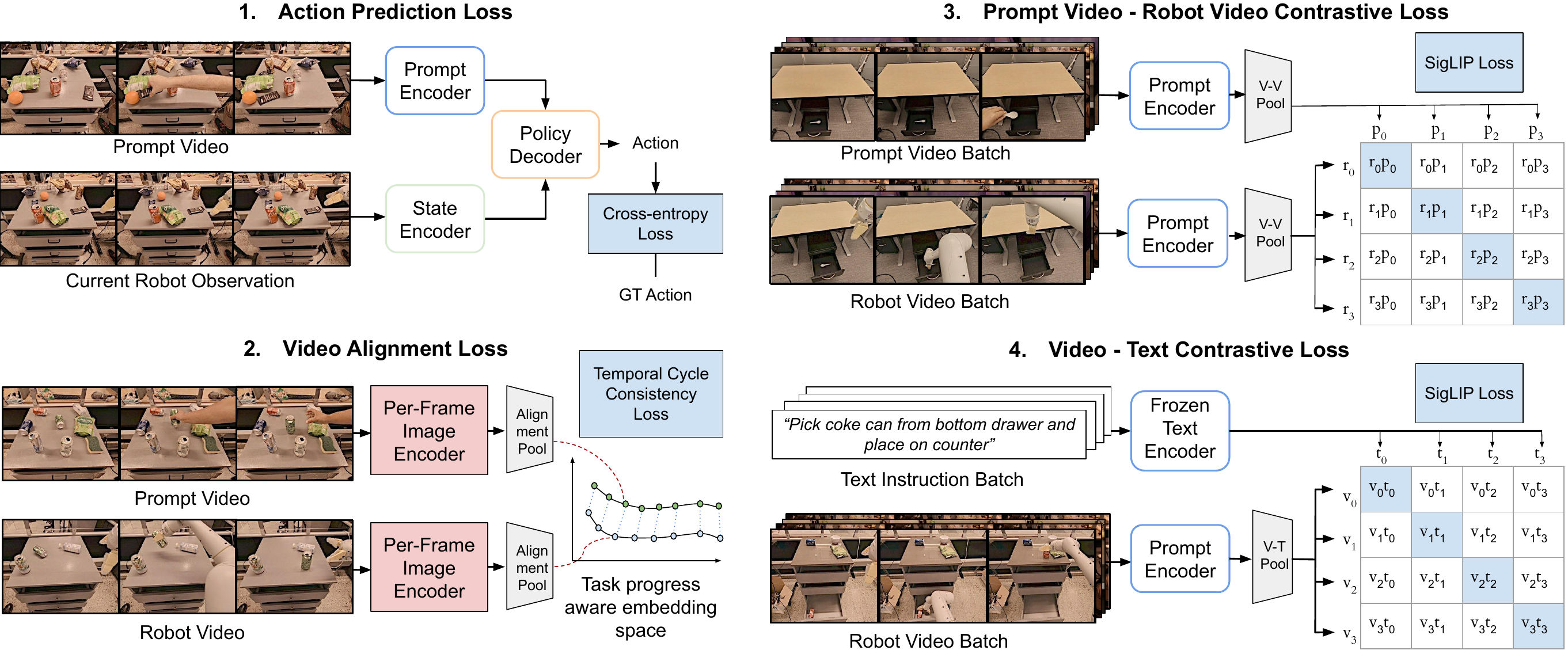}
\end{center}
   \caption{\textbf{Training Setup.} We show all the losses used for training Vid2Robot, particularly how each loss connects to its different modules. Along with (1) the main action prediction loss,
we apply three auxiliary losses: (2) temporal video alignment loss, (3) a contrastive loss between the prompt and robot video performing the same task, and (4) a contrastive loss between a prompt/robot video with the language embedding. More details are in Section~\ref{subsec:training}. }
\label{fig:training-setup}
\end{figure*}

\textit{(1) Action Prediction Loss.}
We train \shortname{} end-to-end with behavior cloning. 
We use a classification loss on actions discretized into $N\!\!=\!\!256$ bins. Given the robot trajectory for performing a task with current visual observations $x_t$, we have the corresponding expert action $a_t$.
The action prediction loss is Cross Entropy between the predicted action and the expert action as: 
$L_{CE}(a_t, \hat{a}_t) = \sum_{\tau} a_t \log \hat{a}_t$.  This loss trains all the model parameters, as shown in Fig~\ref{fig:architecture}.

Although our dataset size is substantial, it is insufficient for training large transformer-based models. To prevent over-fitting on the training set, we add three \textbf{auxiliary losses} to encourage learning features in prompt videos. 

\textit{(2) Video Alignment Loss}: We want to encourage temporal alignment between prompt videos and robot videos performing that show the same task. By aligning prompt videos with the robot videos, we want the image encoder to learn to be invariant to different embodiments, lighting, backgrounds, view angles, and distractor objects while still encoding features relevant to predicting task progress. 

Our choice of loss is the temporal-cycle consistency loss introduced in~\cite{Dwibedi2019-zg}. This loss can encode the task progress when trained on videos of different agents performing the same task~\cite{zakka2022xirl}. This loss is applied on per-frame image embeddings of the prompt $V_p$ and robot $V_r$ videos during training. To apply the loss, we average pool the per-frame embeddings output in spatial dimensions from image encoder $\phi$ and apply a projector head of 2-layer MLP~\cite{chen2020big}. We call this as \textit{alignment pooling layer} $\Phi$ on the per-frame image embeddings, as shown in Fig \ref{fig:training-setup}. For each video $V_i$, this results in a sequence of embeddings $E_i = \{\Phi(v_i^1),\Phi(v_i^2),...\Phi(v_i^{L_i}) \}$, where $L_i$ is the number of frames in the $i^{th}$ video. 
We apply TCC loss on encoding $E_p$ and $E_r$ for prompt and robot video, respectively. 
Intuitively, the TCC loss ensures that the representation of every frame of $E_p$ should correspond to $E_r$ and vice versa. Applying TCC in Vid2Robot involves two steps:
First, we compute soft neighbor of $t^{th}$ frame of $E_p$ ($E_p^t$ in short) in $E_r$ and call it $\widetilde{E_{pr}^t}$. 
\begin{equation}
\widetilde{E_{pr}^t} = \sum_k^{L_r} \alpha_k E_r^k, \quad \mathrm{where} \quad \alpha_k = \frac{e^{-\Vert E_i^t-E_j^k\Vert^2}}{\sum_k^{L_j} e^{-\Vert E_i^t-E_j^k \Vert^2}}
\label{eq:softnncompute}
\end{equation} 

Second, we find the corresponding frame for this newly computed soft-neighbor in $E_p$.  This is called \textit{cycle-back} in~\cite{Dwibedi2019-zg} and it involves similar soft-neighbour computation as in Equation~\ref{eq:softnncompute} to obtain say $\widehat{E_{pr}^t}$, which ideally should be same as $t$, that is, $(\widehat{E_{pr}^t} - t)^2$ should be minimized. 
TCC loss minimizes such mean squared error between all frames for prompt and robot video encodings, and vice-versa, that is,  
\begin{equation}
\begin{aligned}
    L_{TCC}(E_p, E_r) = \sum_{t \in V_p} (\widehat{E_{pr}^t} - t)^2\\
    L_{TCC} = \frac{L_{TCC}(E_p, E_r)+L_{TCC}(E_r, E_p)}{2}
    \end{aligned}
\label{eq:tcc}
\end{equation}

\textit{(3) Prompt-Robot Video Contrastive Loss (VVCL)}: We want to encourage the prompt encodings to learn task semantics from video tokens in a self-supervised manner. While we pair prompt and robot video using natural language, this does not effectively capture the visual similarity of low-level motions like reaching for objects and rotating the robot arm. For this, we apply contrastive loss between the latent features of the robot and the prompt videos.  We use an Attention Pooling layer to merge features from the $N$ prompt tokens to produce a single embedding for each video. We apply the SigLIP~\cite{zhai-siglip-2023} loss between video-video pairs to encourage videos showing the same task, involving similar motions and interacting objects, to be close to each other while being away from other videos in the batch. A batch contains the same number of robot and prompt videos, say $B$. We use the prompt encoder $\psi_p(\phi(\cdot))$ to obtain a batch of full robot video embeddings $Z_{robot}$  and prompt video embeddings $Z_{prompt}$, each of size $B \times d$. We multiply them, $Z_{robot} \cdot Z_{prompt}^T$ to obtain a $B \times B$ matrix. Adding a learnable temperature $\tau$ and bias $b$, we have our logit matrix as $\hat{Y} = (Z_{robot}\cdot Z_{prompt}^T) * \tau + b$. We consider the videos of robot and prompt performing the same task as positives and assign them a label of 1 along the diagonal and -1 for off-diagonal pairs, that is, the label matrix $Y = 2\mathrm{I}_B -1$. SigLIP loss is the negative loglikelihood $\sigma'(Z_1, Z_2) = -\sum \log \sigma(Y \cdot (Z_1\cdot Z_2^T)*t + b)$.
The video-video contrastive loss is as follows:
\begin{equation}
    L_{VVCL} = \sigma'(Z_{prompt}, Z_{robot})
    \label{eq:vvcl}
\end{equation}

\textit{(4) Video-text Contrastive Loss (VTCL)}: We want to encourage a part of the embedding space to be aware of object names and verbs, as shown in the prompt and the robot videos. \replaced{We apply}{Finally, we add} a contrastive loss between prompt tokens produced by the robot video and the text instructions of the task.  A version of this loss has been applied before by BC-Z~\cite{Jang2022-mz} as auxiliary
language regression loss. We use an Attention Pooling layer~\cite{Yu_undated-jc} with one latent query to merge features from the $N$ prompt tokens to produce a single embedding for each video. We retrieve $B$ pairs of video and text embeddings as a batch. Similar to Equation~\ref{eq:vvcl}, we apply SigLIP \cite{zhai-siglip-2023} loss as $L_{VTCL}$ to encourage every video to have similar embeddings to their textual description embeddings, and be different from other text embeddings in the batch.

\vspace{-1.5em}
\begin{equation}
     L_{VTCL} = (\sigma'(Z_{prompt},  Z_{text}) + \sigma'(Z_{robot}, Z_{text}))/2
\end{equation}

Overall, we apply the mean of all four losses for training that is $L = \frac{1}{4} ( L_{CE} + L_{TCC} + L_{VVCL} + L_{VTCL})$.

\subsection{Implementation}
We trained the model (implemented in Jax) for 200K iterations. We use AdamW optimizer with an initial learning rate of 8e-5 using a cosine learning rate schedule with warmup steps 2,000 and a final learning rate of 1e-6. We use 2 Perceiver Resampler layers with 64 latent tokens for both the Prompt and State Resamplers. Both state-prompt encoder and action decoder are 4-layer deep cross-attention transformers. 

%% file: sections/experiments.tex
\section{Experiments}
\label{sec:exp}

\begin{figure*}[t]
\begin{center}
   \includegraphics[width=\linewidth]{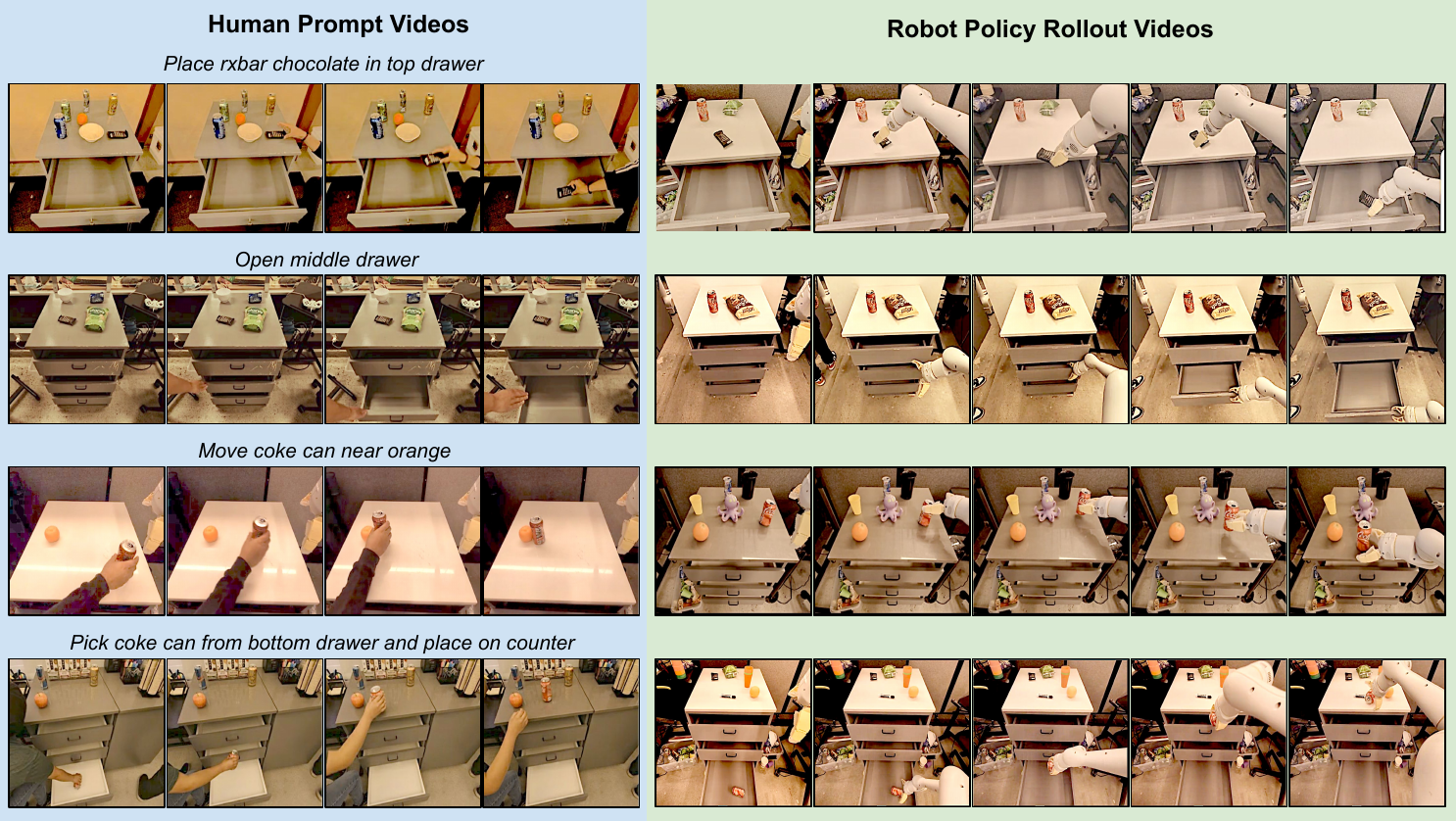}
\end{center}
   \caption{\textbf{Policy Rollouts.} Each row shows a prompt video of a human doing a task on the left, and on the right, we show the corresponding successful robot rollouts using 
   \shortname{}. Note how visually different the prompts are, while the policy rollouts have different lighting and backgrounds, as well as the number and placement of the distractor objects.}
\label{fig:rollout}
\end{figure*}
\input{results/mainresults}

We present results with real robot evaluations for our multi-task video-conditioned policy. 
One of the fundamental questions we tackle in this work is how well robots can imitate humans performing manipulation tasks. Because of differences in embodiments, humans perform manipulation tasks at a different speed and style. We study the effect of using robots and human videos as prompts.

\textit{\textbf{Metrics.}}
A \textit{rollout} as a sequence of actions inferred from the policy and executed on the robot from an initial state observation until the policy terminates or takes the maximum number of steps, whichever is lower. 

We define \textit{success} for a rollout when the policy executes the task instruction as shown in the prompt video. A successful rollout involves correct actions to be taken successively in the environment without any assistance for resets or recovery. 
We ask a human evaluator to observe whether: 
(1) whether the robot \textit{reached} the correct location?, 
(2) \textit{grasped} the correct object?, 
(3) \textit{released} the object at the correct location?, and 
(4) \textit{terminated} the task correctly? 

If the answer to any question is ``no", the answer to subsequent questions is assumed to be ``no". If all questions are answered ``yes", only then the rollout is considered ``successful''.
For each task instruction, we record a few rollouts per policy with different distractor objects, background, and lighting conditions. 
We take the average success recorded across all the rollouts for a task and call it that task's \textit{Success Rate}. We also report aggregated success rate across tasks as \textit{Overall} Success Rate. We also analyze the partial success across the four milestones in Section~\ref{subsec:tasksuccess}.

We looked into automating success detection with concrete criteria or decision rules for evaluation but found that this automation can overlook the \textit{process} of performing the task. Consider the task of ``knocking the water bottle over". 
\textit{Case 1:} The robot successfully grasps the bottle, turns it, and places it on the table. 
\textit{Case 2:} The robot fails to grasp, pinches the bottle away, and lands in a knocked-down orientation. While Case 1 is desired and expected behavior according to the training data, Case 2 is a failure as it is unintended. With rule-based verification of the final state, we would have deemed both Case 1 and 2 successful. With human evaluators, we focus on the entire process of achieving the task instead of the final state.

\textit{\textbf{Setup.}} We evaluate the policies by varying the object placement, lighting conditions, background surfaces, and distractor objects.
When evaluating a set of policies, we ensure comparable initial object configurations for rollouts. We randomize the initial state only after all policies' rollouts have been performed.
For all rollouts, we sample prompt videos not seen during training. 
In all the experiments, we use a mobile manipulator, the
\href{www.everydayrobots.com}{Google Robot}. It has an ego-centric camera view, a single arm with seven degrees of freedom, and a two-fingered soft gripper. Refer to \cite{Brohan2022-ta} for more details.

\textit{\textbf{Baselines.}}
We compare our model with BC-Z \cite{Jang2022-mz}, a video-conditioned policy using a ResNet-18 encoder. BC-Z \citep{jang2022bc} processes demonstration-observation pairs via a FiLM \citep{perez2018film} conditioned ResNet encoder and feds into a ResNet-based policy network to predict robot actions. We use the same data to train the BC-Z and Vid2Robot for a fair comparison. BC-Z does not have a terminate action, so we run these rollouts for a fixed maximum number of steps.

\textit{\textbf{Key Questions and Results.}}
We address the following questions in this work: 
(1) What is the success rate gap due to prompt embodiment (robot vs. human) across tasks?  (\S\ref{subsubsec:gap}) (2) How do video-conditioned policies perform with unseen task videos? (Fig~\ref{fig:rollout}, \S\ref{subsubsec:unseenvideo}) (3) Is Vid2Robot's overall success significantly better than BC-Z baseline? (\S\ref{subsec:statsig}) (4) Can learned motion representations handle out-of-distribution object interactions?  (\S\ref{subsec:verbsuccess})

\subsection{Task-based success}
\label{subsec:tasksuccess}
We compare our \shortname{} model and baseline BC-Z with robot- and human-performed prompt videos in Table~\ref{tab:mainresults}. We train both \shortname{} and BC-Z  on the same data mixture containing robot-robot and human-robot paired data. Prompt videos cover a subset of the training tasks. However, the videos are unseen by the models. In this evaluation, we investigate each model's ability to infer the task specification from the prompt video as well as the current observed state of the robot.

To test the model's capabilities in different settings on real robots, we assess rollouts on the following nine tasks:  
    `knock water bottle over',
    `move rxbar chocolate near coke can',
    `move green jalapeno chip bag near coke can',
    `pick green rice chip bag',
    `place coke can upright',
    `pick coke can from bottom drawer and place on counter',
    `open middle drawer',
    `close middle drawer', and
    `place apple into top drawer'. 
    
We ask four evaluators to carry out two rollouts per task for a prompt video dataset and policy setting (a row in Table~\ref{tab:mainresults}). Overall, we have eight trials per task to evaluate a policy's task success rate. We report an overall success rate per row over nine tasks with eight trials per task, that is, $9\!\times\!8\!=\!72$ trials. In total, we required $72\!\times\!4\!=\!288$ rollouts for Table~\ref{tab:mainresults}.

\vspace{1em}
\subsubsection{What is the gap in success rate due to embodiment difference in prompt videos?}
\label{subsubsec:gap}
\begin{figure*}[t]
    \centering
    \includegraphics[width=\linewidth]{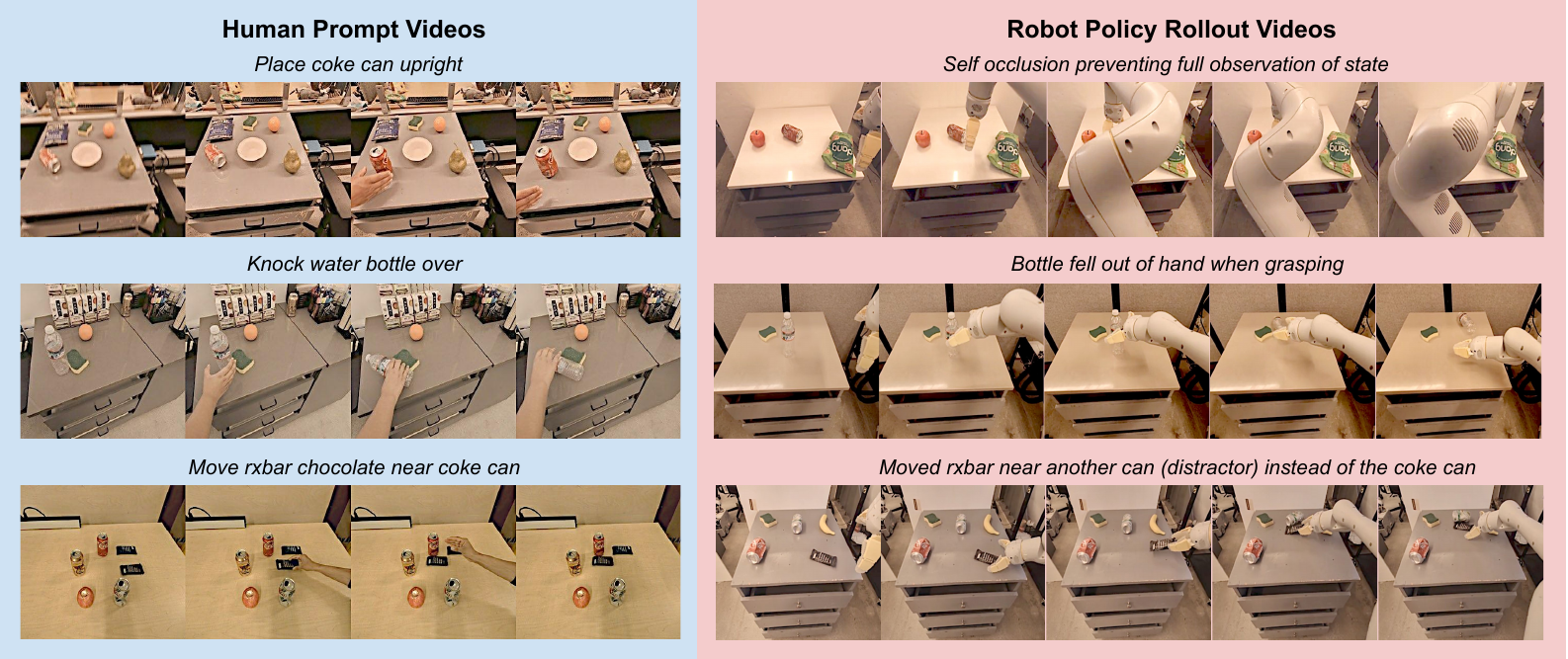}
    \caption{\textbf{Failure analysis with policy rollouts.} (Top) Policy predicts gripper pose and depends on the IK solver to move the arm. Sometimes, the IK solution can block the robot's camera view. (Middle) Grasping failures happen, especially with transparent and deformable objects. (Bottom) Distractor objects and differences in lighting and background may cause recognition errors, where policy might perform the correct motion but with an incorrect object(s).}
    \label{fig:policyfailures}
\end{figure*}

When prompted with robot and human videos, we compare our model with BC-Z, which is a strong baseline for our comparisons.  
The overall success rate of our model \shortname{} outperforms BC-Z for Human prompt videos by 20\%, and is comparable for Robot prompt videos. Note that there is an order of magnitude more training samples for robot trajectories than human videos in our training mixture. Hence, there isn't a significant gap in performance for robot prompt videos. Our model outperforms BC-Z in most tas for human prompt video, showing that \shortname{} captures the task semantics from prompt videos better than the baseline. 
Our model outperforms in tasks like picking something from a drawer, placing it on the counter, and opening/closing drawers by a large margin. 
The most challenging task is \textit{placing upright} and \textit{knocking over}. We analyze the failure reasons in \S\ref{sec:limitations} Fig~\ref{fig:policyfailures}. 

\vspace{1em}
\subsubsection{How well do video-conditioned policies perform when shown a task in an unseen video?}
\label{subsubsec:unseenvideo}
In addition to marking a rollout as a success, we recorded partial success annotations per rollout.
In Fig~\ref{fig:partialsuccess}, we observe that our model \textit{reaches} to the correct object 78\%, about 8\% more than baseline. The policies sometimes fail to get the correct object and go towards a distractor instead. Next, \textit{grasping} errors happen, particularly with small and deformable objects and collision-prone areas like drawer handles or counter's edges. Here, our model (65\%) outperforms BC-Z (45\%) by a large margin of 20\% — a successful grasp is often the most challenging part of a rollout and crucial for success.  
After grasping, most tasks require \textit{releasing} at a correct location. Both models slightly drop in success rate due to incorrect \textit{release} during the rollouts. While BC-Z runs for a fixed number of steps, our policy \shortname{} predicts when to terminate. 
We observe that the rate of \textit{release} and \textit{terminate} is almost identical, about 57\% for our model, which implies that after releasing at the correct location, \shortname{} mostly terminates successfully.   
\input{results/partialsuccess}

\subsection{Tasks with More Rollouts}
\label{subsec:statsig}
To comment on the statistical significance of our results, we conducted more trials while limiting the evaluation to two tasks, namely `place coke can upright' and `close middle drawer' for real robot policy evaluations, and reported mean success rate with confidence intervals. In total, we conducted 314 real robot rollouts for results reported in Table \ref{tab:statsig}. 
 
\input{results/statsig}

\subsection{Cross-object motion transfer} 
\label{subsec:verbsuccess}

 \begin{figure}[t]
\begin{center}
   \includegraphics[width=\linewidth]{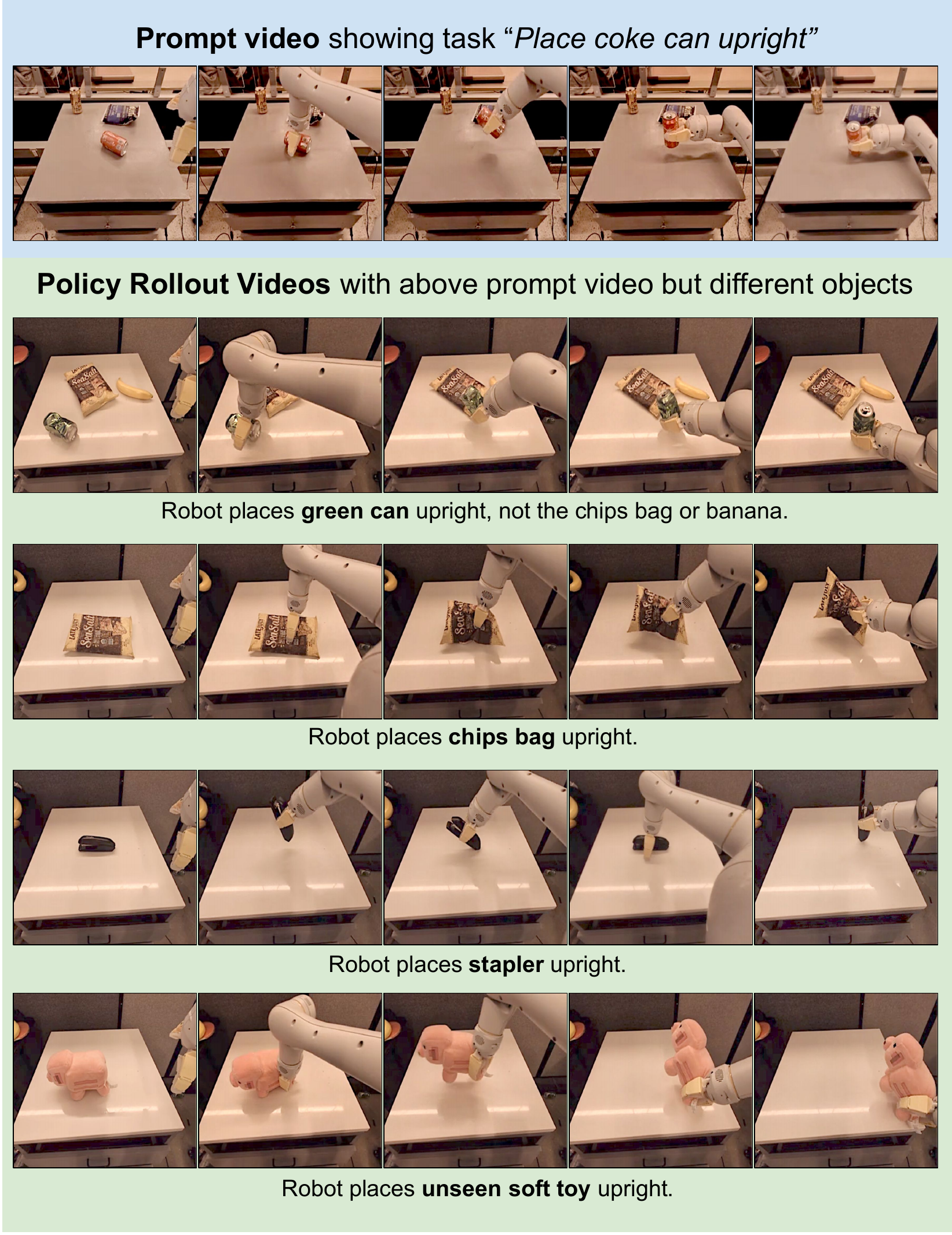}
\end{center}
   \caption{\textbf{Qualitative results for \textit{cross-object motion transfer}.} (Top-blue) Prompt video of \textit{placing coke can upright}; (Green) Policy rollouts with a \textit{green can}, \textit{chips bag}, \textit{stapler} and a \textit{soft toy} in front of the robot. \shortname{} infers the motion of \textit{place upright} in the prompt video and applies it to other objects. The policy adheres to the prompt video by picking the green can instead of the chips bag or banana.}
\vspace{-1.5em}
\label{fig:qualitative}
\end{figure}

We trained our \shortname{} and baseline with paired videos as discussed in Section~\ref{subsec:datasets}. Due to the pairing, the training data included only those scenarios where the interaction object shown in the prompt is present in the current robot observations. But \textit{what if we provided a prompt video of one object and tested on other objects? Does it make the same motion as shown in the prompt video?} 
Interestingly, we found our model to perform learned manipulation actions on objects not seen in the prompt video. We call this emergent behavior as \textit{cross-object motion transfer}. 

We compare \shortname{} with BC-Z for cross-object motion transfer ability with five prompt videos, namely, 
    `knock water bottle over',
    `pick green rice chip bag',
    `place coke can upright',
    `pick coke can from bottom drawer and place on counter', and
    `place apple into top drawer'. 
We evaluate each case of a prompt video by placing unrelated objects in robot's initial observation. The objects used for evaluation are \textit{`orange', `green can', `chips bag', `banana', `pink piggy soft toy', `wrist watch'}. We selected objects with different shapes, sizes, and deformability to evaluate situations requiring different grasps for success.

The evaluation setup is similar to Section~\ref{subsec:tasksuccess}. Here, the evaluator sets up one of the objects for a task and records rollouts for each model. We compare two models on five tasks with six objects, so every evaluator runs $2\!\times\!5\!\times\!6\!=\!60$ rollouts.
We repeat the evaluation with four raters, thus reporting results in Table~\ref{tab:verbtransfer} on $4\!\times\!60 = 240$ rollouts. 

In Fig~\ref{fig:qualitative}, we show the above experimental setup qualitatively. We use a prompt video to `place coke can upright' and observe that the policy can transfer the action of `placing upright' to objects, like a green can, a chips bag, a stapler, and a soft toy. The policy shows an implicit notion of learned pragmatics, by selecting green can over other objects.

In Table~\ref{tab:verbtransfer}, we observe that BC-Z is often unable to complete the tasks when testing \replaced{cross-object}{cross=object} motion transfer. In contrast, our model (34\%) performs better than BC-Z (17\%) in this setting and performs the motion indicated in the prompt video. 
Our model is comparable to BC-Z with a 45\%  success rate on \textit{picking} out-of-distribution objects. More importantly, tasks involving placing into drawers demonstrate significant improvement ($29\% \rightarrow 54\%$). For specific tasks like \textit{picking from drawers}, \textit{placing on counters}, and \textit{knocking over}, \shortname{} completes the task $25\%-29\%$ of the time, whereas BC-Z is unable to perform.
 
\input{results/verbtransfer}

\subsection{Ablations}
\label{subsec:ablations}

We analyze our proposed approach to understand the following: (a) Can the policy learn possible interactions with the environment from state observations alone, without needing prompt videos? (b) How do the auxiliary loss functions impact the model's performance?  

\subsubsection{What is the impact of the prompt for task inference?}
First, we motivate the importance of prompt videos with an example. Consider a Coke can and other objects on the countertop as the robot's state observation. If a Coke can exists in the robot's view, it is hard to infer whether the task is to ``pick a Coke can”, ``move a Coke can close to a chocolate bar”, ``move a Coke can near an orange can,” or ``knock over Coke can”. Furthermore, once the robot starts taking action, it can end up in new states, like being close to rxbar,  which make it especially difficult to predict tasks from the current state only. 
Below, we empirically measure the success rate of a policy that does not have access to a suitable prompt video. 

We evaluated the ``no-prompt'' case, in which both models see blank frames as input prompt videos. In this setup, we evaluated three tasks. For each task, we rolled out the policy 20 times. 
In total, we ran $2\times20\times3=120$ actual robot rollouts for this experiment. Here, the success rate is 23\% for Vid2Robot and  5\% for BC-Z over 20 rollouts per task per policy. 
We find that performance improves when we condition the policies on the prompt videos. The success rate improves from $5\%$ to $52.6\%$ for BC-Z and from $23\%$ to $54.6\%$ for Vid2Robot. (Refer Table~\ref{tab:mainresults}). This experiment underlines the importance of prompt videos for task success.

\subsubsection{What is the role of auxilliary losses?}
Second, we analyze the role of additional loss functions in the overall success rate. In Section~\ref{subsec:training}, we presented action prediction loss and three auxiliary losses. 
We investigate the impact of (1) not using any auxiliary loss and (2) adding auxiliary language loss. 
We consider the tasks similar to those described in Section~\ref{subsec:tasksuccess}, 9 tasks for evaluating each policy. 
We have 3 model variants: the original \shortname{}, the one without video-text contrastive loss (CL), and the one with only action prediction loss. 
We ask three human evaluators to run a model variant for two rollouts each. In total, we report results with  $3\!\times\!3\!\times\!9\!\times\!2\!=\!162$ rollouts in Fig~\ref{fig:lossablations}. The error bars show the standard deviation for success.

\textit{What is the impact of not using any auxiliary loss?}
We observe that the performance of our model (61\%) improves significantly by enforcing representation constraints through auxiliary losses, compared to using only action prediction loss (45\%). It highlights the importance of the proposed auxiliary losses in Section~\ref{subsec:training}.

\textit{What is the impact of the auxiliary language loss?} 
BC-Z proposed to use language representations to improve video representations for conditioning the policy. We compare our policy with another variant trained with all losses but the Video-Text CL. We observe only a borderline improvement of 1-2\% in the success rate when using language loss. 
This implies that video alignment and video contrastive loss contribute significantly towards performance improvement.

\input{results/ablations}

%% file: results/mainresults.tex
\begin{table*}[t]
    \centering
        \caption{Task Success Rate for Robot and Human prompts.}
    \begin{tabular}{clrrrrrrrrr}
    \toprule
Prompter     & Model    & \textit{pick}    & \textit{pick-place on}    & \textit{place into}    & \textit{open}    & \textit{close}    & \textit{move near}    & \textit{knock over}    & \textit{place upright}    & Overall \\
\midrule
\multirow{2}{*}{Robot}     & BC-Z    & 75.0\%    & 50.0\%    & \textbf{61.5}\%    & 16.7\%    & 66.7\%    & \textbf{44.0}\%    & \textbf{58.3}\%    & \textbf{50.0}\%    &  52.6\% \\
  & Vid2Robot    &75.0\%    &\textbf{58.8\% }   &50.0\%    &\textbf{91.7\%}    &\textbf{100.0\%   } &33.3\%    &41.7\%    &16.7\%    &  \textbf{54.9\%} \\
\midrule
\multirow{2}{*}{Human}     & BC-Z    & 50.0\%    & 12.5\%    & 12.5\%    & 0.0\%    & 50.0\%    & 43.8\%    & 12.5\%    & \textbf{50.0}\%    &  30.6\% \\
  & Vid2Robot    & \textbf{100.0\%}    & \textbf{50.0\%}    & \textbf{50.0\%}    & \textbf{62.5\%}    & \textbf{87.5\%}    & \textbf{43.8\% }   &\textbf{ 25.0\%}    & 12.5\%    &  \textbf{52.8\%} \\
\bottomrule
\end{tabular}
    \label{tab:mainresults}
\end{table*}

%% file: results/partialsuccess.tex
\begin{figure}[t]
\begin{center}
  \includegraphics[width=\linewidth]{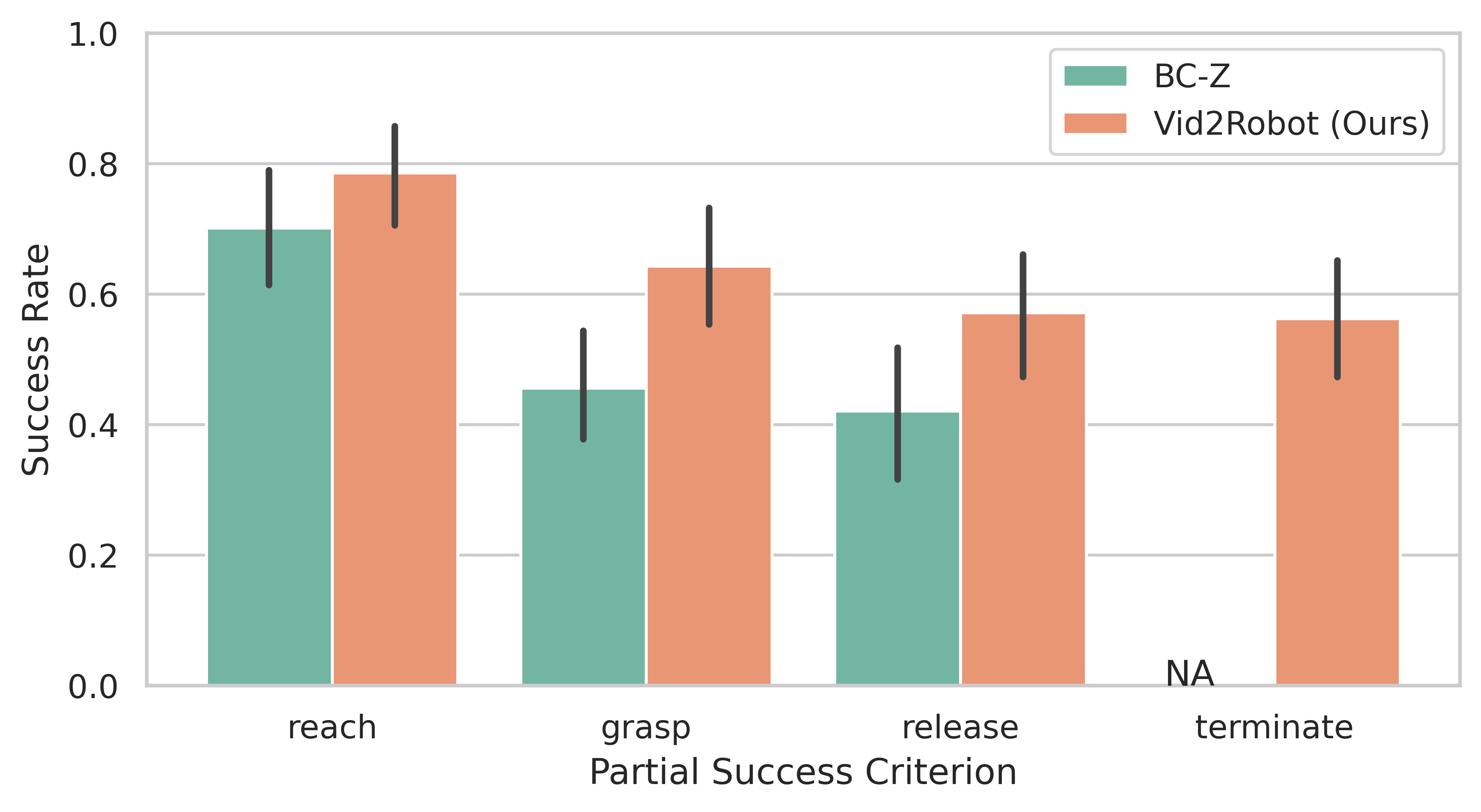}
\end{center}
  \caption{\textbf{Partial Success Rate for BC-Z and Vid2Robot. }Our policy \shortname{} outperforms BC-Z in terms of \textit{reaching} the correct object, \textit{grasping} it, \textit{releasing} it at the correct location and then \textit{terminating }the episode correctly. Note that BC-Z does not have terminate control.}
\label{fig:partialsuccess}
\end{figure}

%% file: results/statsig.tex
\begin{table}[t]
\centering
\caption{Real Robot Evaluation of  Vid2Robot and BC-Z with more trials to ascertain the statistical significance of the results.}
\begin{tabular}{@{}lrrr@{}}
\toprule
Model     &  place coke can upright & close middle drawer & Overall\\ \midrule
BC-Z       & 19.4 $\pm$ 9.5\%   &    39.2 $\pm$ 10.8\%                                                               & 30.2 $\pm$ 7.1\%                          \\
Vid2Robot                                                          &  \textbf{39.1} $\pm$ 9.9\%   &        \textbf{48.7} $\pm$ 11.2\%                                                            & \textbf{43.4} $\pm$ 7.2\%      \\
\bottomrule
\end{tabular}
\label{tab:statsig}
\end{table}

%% file: results/verbtransfer.tex
\begin{table}[t]
    \centering
    \caption{Cross-object motion transfer success.}
    \begin{tabular}{@{}lcccccc@{}}
\toprule
 &   & \textit{pick-}   & \textit{place}   & \textit{place}   & \textit{knock}   &  \\
Model    & \textit{pick}   & \textit{place on}   & \textit{into}   & \textit{upright}   & \textit{over}   & Overall \\
\midrule
BC-Z     & 45.8\%    & \phantom{0}0.0\%    & 29.2\%    & 12.5\%    & \phantom{0}0.0\%    & 17.5\% \\
\shortname{}   & 45.8\%    & \textbf{25.0\%}    & \textbf{54.2\% }   &\textbf{ 16.7\% }   & \textbf{29.2\%}   & \textbf{34.2\%} \\
\bottomrule
    \end{tabular}
    \label{tab:verbtransfer}
\end{table}

%% file: results/ablations.tex
\begin{figure}[t]
\begin{center}
  \includegraphics[width=\linewidth]{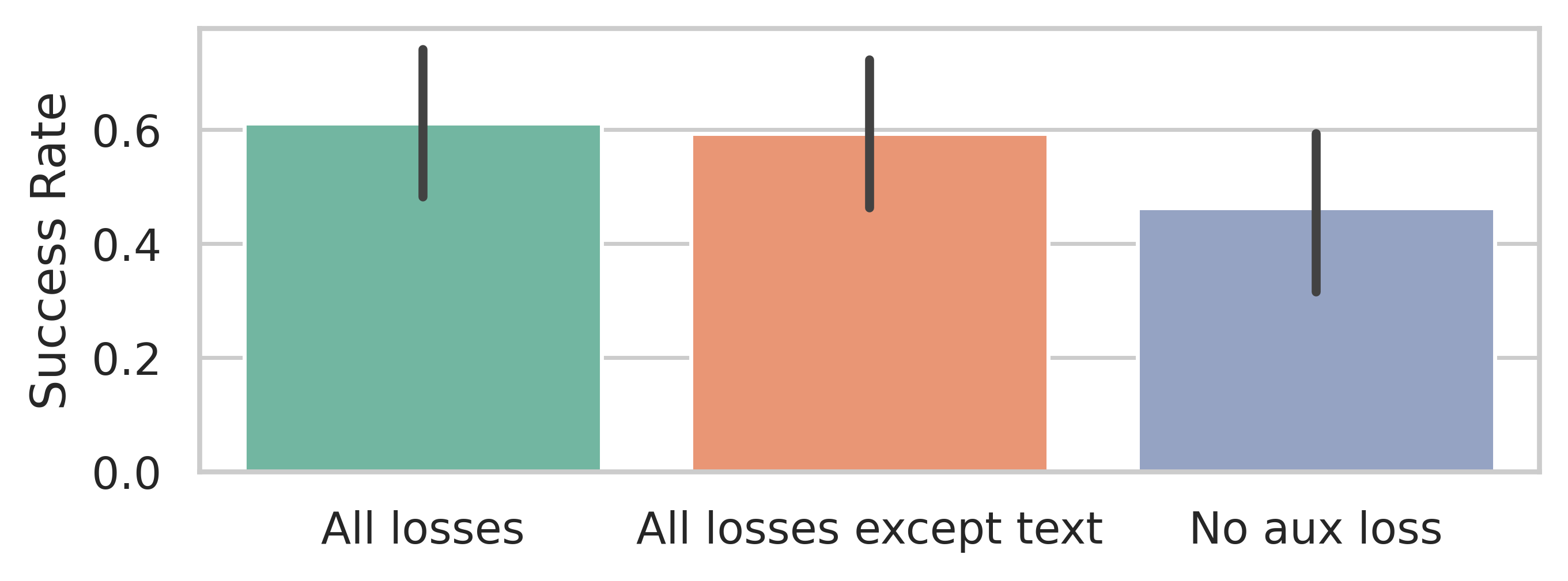}
\end{center}
  \caption{\textbf{Ablation for auxilliary losses used in \shortname{}.} We compare our proposed approach that has all auxiliary losses (green, left) with a variant without language contrastive loss that was originally proposed in BC-Z (orange, middle) and a version with no auxilliary losses (blue, right). More details in (\S \ref{subsec:ablations})}
\label{fig:lossablations}
\end{figure}

%% file: sections/relatedwork.tex
\section{Related Work}
\label{sec:relatedwork}

\subsection{Task Specifications for Robots}
The development of general-purpose robots hinges on effectively grounding task specifications. 
Videos are a dense source of information that provides information about what to do and how to do it in the physical world. Recent works have used videos for task specification \cite{Bahl2023-ut, Jiang2022-wa, Shah2023}. Another line of work uses videos to learn world models to predict future visual observations \cite{mendonca2023structured, Liu2018-ak, Chane-Sane2021-jq, Nair2018-nv, du2023learning}. 

Recall our example of ``open drawer”, ``open cabinet”, and ``open jar” in the \S{\ref{sec:intro}}. Video-conditioned policies like Vid2Robot are capable of doing these tasks because these policies identify the tasks of "open jar" and "open drawer" from visuals, unlike language-conditioned which have the same embedding of the verb `open' for each task. Note that language is not an input to Vid2Robot. Therefore, the verb does not directly influence the action; this is a critical difference between Vid2Robot and existing language-conditioned robot policies.
While language \cite{Yenamandra2023-rn, Brohan2022-ta, Open_X-Embodiment_Collaboration2023-ee, Parashar2023-cy}, final goal images \cite{Krantz2023-qg, Bousmalis2023RoboCatAS}, and others like hand-drawn inputs \cite{Stone2023-yh} have been proposed as means for task specification, learning from prompt videos is complementary to these approaches and inevitable for rapid adaptation of trained polices to perform new manipulation skills at deployment.

\subsection{Learning from Human Demonstrations} 
As videos of humans performing various tasks proliferate on the internet, several works aim to address how to leverage this information for robot learning. 
The difference in robot versus human embodiment poses a significant challenge, for which existing approaches range from translating the image of a human into the robot \citep{Smith2019-gg} to inpainting for agent-agnostic representations \citep{bahl2022human}. 
Many prior works propose to leverage off-the-shelf models for hand pose estimation and contact tracking \citep{bahl2023affordances, Dessalene2023-sd, Petrik2021-xu}, object-centric representations \cite{PirkOnlineObjects2019,pmlr-v205-jain23a}, as well as reward functions for reinforcement learning \cite{bahl2022human, Ma2022-il, Smith2019-gg}. 

XIRL \cite{zakka2022xirl}, GraphIRL\cite{kumar2022graph} and other RL approaches \cite{Schmeckpeper2020-wo, Peng2018-rc} take a lot of time and manual effort for resetting scenes during the policy learning phase, limiting their applicability to real robots. Furthermore, RL often leads to unsafe situations with real-world robots. We compare Vid2Robot to another end-to-end behavior cloning method, BC-Z, that has been shown to scale to multiple tasks. 
Other methods \citep{nair2022r3m, Xiao2022-by, bahl2023affordances} cast this problem into visual representation learning to accelerate learning of downstream motor control tasks. %
While these modular learning solutions work well in limited datasets, they are prone to compounding errors in each component and are not efficiently scalable. 
End-to-end training approaches for goal-conditioned imitation learning \cite{dasari2021transformers, Sharma2019-lw, Groth2021-ri, Ding2019-ac} are also largely limited in simulation and hindered by sim-to-real gap. 
In contrast, we tackle this as an end-to-end large multi-task learning from human videos with real robot evaluations.  

\subsection{Imitation via Paired Demonstrations}
Our setup of paired prompt videos and robot trajectory is most similar to the One-Shot Visual Imitation literature. Many prior works assume access to pairs, where the first video demonstrates the task, and the second video shows the agent's visual observations. 
Some of the early works \citep{duan2017one} proposed training a demonstration network via temporal convolution and neighborhood attention to condition a manipulation policy network. In more recent approaches like  \citep{dasari2021transformers, mandi2022towards, pmlr-v205-jain23a}, paired demonstrations and observations are used to train a transformer policy, often with additional constraints like inverse dynamics prediction\citep{dasari2021transformers} or contrastive representation learning \citep{mandi2022towards}. MimicPlay~\cite{wang2023mimicplay} proposes hierarchical learning framework using human and robot teleoperated demos. However, evaluating these approaches is usually limited to a specific set of tasks, without any changes in background between the two videos.

BC-Z \citep{Jang2022-mz} is most similar to our work, which reports real robot evaluations. While our training setup has similarities with BC-Z, our model \shortname{} couples large image encoders, cross-attention layers, and contrastive auxiliary losses to learn a manipulation policy that imitates a human showing a task. Recent approaches for self-supervised skill discovery like XSkill~\cite{xu2023xskill} learn skills from unpaired human and robot videos, while our approach uses text descriptions of the task to pair them explicitly.  The paired human and robot videos contain different backgrounds, lighting, and object arrangements, thereby training the visual representations to be invariant to these settings, and focus on the task semantics instead.

%% file: sections/limitation.tex
\section{Limitations and Future Directions}
\label{sec:limitations}

In Sec \ref{sec:exp}, we show that our approach has improved over previous work but there is a gap in performance for video-conditioned policies. 
Below we discuss the limitations of our work and provide insights for the future.

First, we qualitatively investigate some reasons for the failure of a policy rollout. In Fig~\ref{fig:policyfailures}, we illustrate and explain three examples of how self-occlusion, grasping errors, and the presence of distractors can lead to failure during any rollout. 

Second, we observe a significant drop in the grasping success in Figure \ref{fig:partialsuccess}. While we use robot camera observation to estimate the state and implicitly learn depth estimation, it is often incomplete when \deleted{there is } occlusion or the robot gripper is out of camera view. Enhancing the state information with multimodal sensor fusion may improve the grasp success rate. 

Third, we consider carefully collected short task instruction demonstrations from three different sources as shown in Section \ref{subsec:datasets}, all of which are 5 to 20-second videos. To test our models on long-horizon demonstrations or `in-the-wild' videos online, we need effective pairing strategies for videos and a few corresponding robot trajectories to train the policy.

%% file: sections/conclusion.tex
\section{Conclusion} 
\label{sec:conclusion}

We present Vid2Robot, an end-to-end video-conditioned robot policy. Our proposed system trains on paired videos such that both videos demonstrate the same task but differ in diverse settings of lighting, background, and distractor objects. We use cross-attention (i) to learn the joint latent representations from prompt and state encodings and then (ii) to decode the action. We train the entire model for action prediction with cross-entropy loss and three auxiliary losses that encourage learning of generalizable latent representations to infer tasks directly from raw pixels to suitable actions. Vid2Robot outperforms BC-Z by over $\sim$20\% when prompted with human videos. Further, Vid2Robot outperforms BC-Z by $\sim$17\% for cross-object motion transfer; that is, if the prompt video didn't have the exact object as the object the robot is manipulating now, the model still produces valid actions for the same verb but different objects. 
Cross-object motion transfer is a promising direction for further extending pragmatic transfer learning of motion to new objects. 
We hope Vid2Robot enables bootstrapping data collection and human-robot interaction with rapid adaptation to new skills. 

%% file: sections/acknowledgements.tex
\section*{Acknowledgements}
We would like to thank Yansong Pang, Grecia Salazar, Utsav Malla, Deeksha Manjunath, Jornell Quiambao, Sarah Nguyen, Sangeetha Ramesh, Tran Pham, Samuel Wan, Tomas Jackson, Jodilyn Peralta, Celeste Barajas, Elio Prado, Rochelle Dela Cruz, Alex Luong and Krista Reymann for supporting data collection via teleoperation. Special thanks to Jornell Quiambao, Grecia Salazar, Utsav Malla, Deeksha Manjunath,  Sarah Nguyen, Sangeetha Ramesh, and Jaspiar Singh for evaluations on robot;   Michael Ahn, Anthony Brohan and Keerthana Gopalakrishnan for policy evaluation infrastructure;
Suneel Belkhale,  Dorsa Sadigh, Chelsea Finn, and Sergey Levine for helpful discussions; Jonathan Tompson, and Vincent Vanhouke for thorough feedback on the writing.
This work was performed by the first author as an intern at Google DeepMind Robotics.
This work was also supported is partially funded by an unrestricted gift from Google and by the Defense Advanced Research Projects Agency (DARPA) under Agreement No. HR00112490375.

%% file: sections/appendix.tex
In our video provided in the supplementary materials, we show several examples of successful rollouts, and model architecture with prompt video and robot observations. We provide specific examples of task-based success, and cross-object motion transfer. Additionally, we provide qualitative results on long-horizon task composition and rollouts for tasks that are rare in our training dataset. Finally, we provide rollouts of noted failures like case of self-occlusion, grasping errors and ambiguous interpretation of the task in the prompt video.

\section{Training Details}

We will release the model code and trained checkpoints. In Table~\ref{tab:archiecture}, we present the detailed version of the \shortname model architecture. In Table~\ref{tab:hyperparams}, we present the hyperparameters used while training. We also add data augmentations to the videos while training as per the settings in Table~\ref{tab:dataaug}.  We normalize the videos 
as required for the pre-trained ViT model. 

\begin{table*}[hbt!]
\centering
\caption{\textbf{Detailed Architecture of \shortname.}  }
\renewcommand{\arraystretch}{1.50}
\begin{tabular}{l|l|c|l}
\toprule
\textbf{Module} & \textbf{Layer} & \textbf{Output Size} & \textbf{Notes} \\ 
\midrule
\hline
\multirow{4}{*}{\textbf{Prompt Encoder}}  & Input  & 16$\times$224$\times$224$\times$3 & Reference Video of 16 frames\\
\cline{2-4} & Image Encoder & 16$\times$196$\times$768 & Each frame passes through ViT-B/16\\
\cline{2-4} & \multirow{1}{*}{Reshape} & 3136$\times$768  & Reshapes all space-time tokens to be in one dimension\\
 \cline{2-4} & \multirow{1}{*}{Perceiver Resampler} & 64$\times$768  & 2 layers with 768 dims and 12 attention heads\\
 \hline

\multirow{4}{*}{\textbf{State Encoder}}  & Input  & 8$\times$224$\times$224$\times$3 & Robot Observation Video of 8 frames\\
\cline{2-4} & Image Encoder & 8$\times$196$\times$768 & Each frame passes through ViT-B/16\\
\cline{2-4} & \multirow{1}{*}{Reshape} & 1568$\times$768  & Reshapes all space-time tokens to be in one dimension\\
 \cline{2-4} & \multirow{1}{*}{Perceiver Resampler} & 64$\times$768  & 2 layers with 768 dims and 12 attention heads\\
 \hline

\multirow{2}{*}{\textbf{State Prompt Encoder}} & Cross-attention & 64$\times$768& 4 layers of cross-attention with 768 dim and 8 attention heads\\
  & Transformer Layer &  & Uses prompt tokens as keys and state tokens as queries\\

\hline

\multirow{2}{*}{\textbf{Action Decoder}} & Cross-attention & 11$\times$768& 4 layers of cross-attention with 768 dim and 8 attention heads\\
  & Transformer Layer &  & Uses prompt-aware state tokens as keys and action position embeddings as queries\\

\hline

\multirow{1}{*}{\textbf{Action Head}} & Linear & 11$\times$256& Projection of action decoder output to 256 bins\\
\bottomrule
\end{tabular}
\vspace{1em}

\label{tab:archiecture}
\end{table*}


\begin{table}[hbt!]
\centering
\caption{\textbf{Hyperparameters for \shortname.}  }
\renewcommand{\arraystretch}{1.50}
\begin{tabular}{l|c}
\toprule
Hyperparameters & Values \\ \midrule
\hline
Batch size & 2048 \\
Learning rate   & 8e-5  \\
Optimizer       & AdamW  \\
Num training steps & 200,000 \\
Warmup steps    & 2000   \\
Image size & 224  \\
Num prompt frames & 16 \\
Num robot frames & 8 \\
Prediction Horizon & 4 \\
\bottomrule
\end{tabular}

\label{tab:hyperparams}
\end{table}

\begin{table}[hbt!]
\centering
\caption{\textbf{Data augmentation for training \shortname.}  }
\renewcommand{\arraystretch}{1.50}
\begin{tabular}{l|c}
\toprule
Hyperparameters & Ranges \\ \midrule
\hline
Height crop range & (0.95, 1.) \\
Width crop range & (0.95, 1.) \\
Brightness range & (0.9, 1.1) \\
Contrast range & (0.8, 1.2) \\
Hue range & (0, 0.03) \\
Saturation range & (0.8, 1.2) \\
\bottomrule
\end{tabular}

\label{tab:dataaug}
\end{table}

\section{Dataset}

We created the dataset mixture from the three sources. 
We collected 120k robot trajectories, 5k human trajectories, and 5k co-located human and robot trajectories. 
To create pairs, we sampled 3 videos as prompts per robot trajectory. This gives 360k pairs for Robot-Robot, about 15k pairs for Hindsight Human-Robot and 5k pairs for Co-located Human-Robot datasets.
To ensure that all pairs are approximately sampled equally, we create the mixture proportions as 90\% Robot-Robot, 5\% Hindsight Human-Robot and 5\% Co-located Human-Robot.

\section{Qualitative Results on Rare Tasks}

\begin{figure*}
    \centering
    \includegraphics[width=\linewidth]{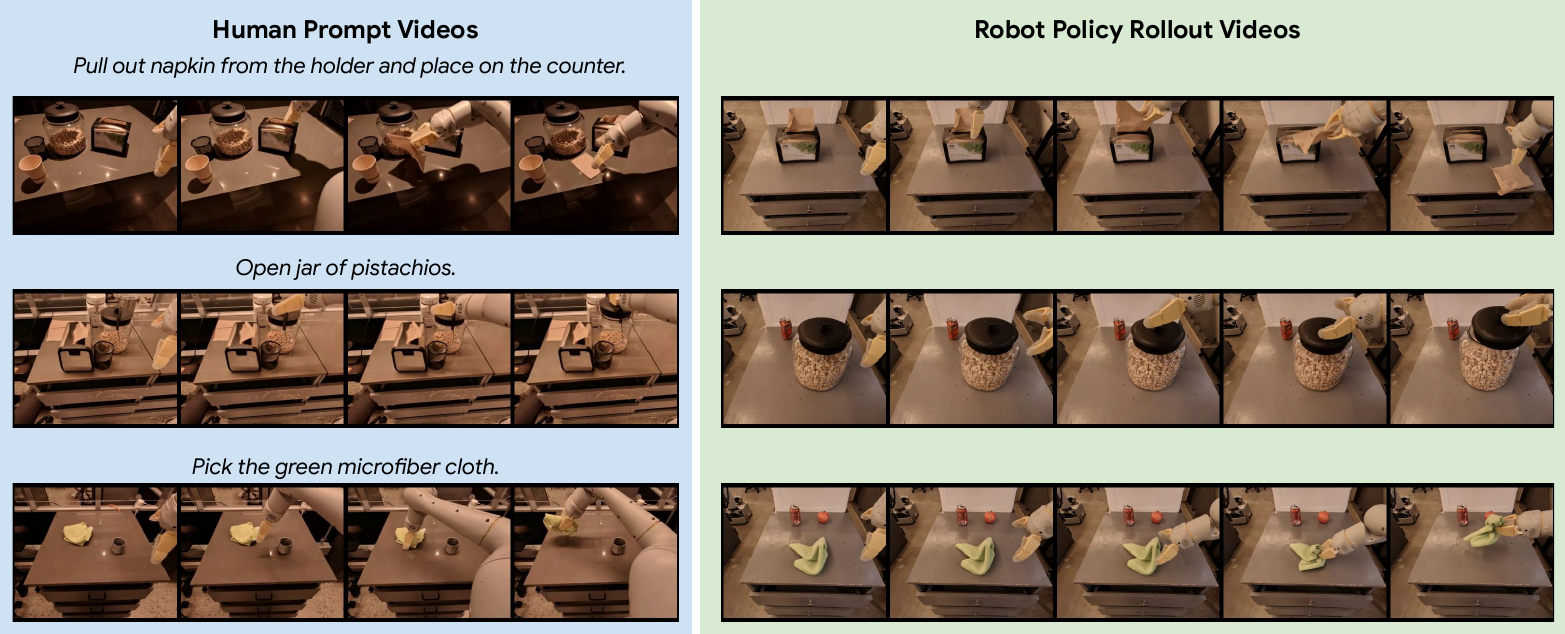}
    \caption{Qualitative results on rare tasks.}
    \label{fig:rare_settings}
\end{figure*}

We also test \shortname on some tasks which are rare in the training set and find that video-conditioned models are also able to complete them. Some examples of these tasks are: \textit{opening glass jar, picking up green micro-fiber cloth and pulling out napkin}. We show examples in Fig.~\ref{fig:rare_settings}. For the video version of these results please take a look at the supplementary video.

\section{Hardware setup}

In all the experiments in this paper, we use mobile manipulators.
These robots have a 7 degree-of-freedom arm, a two-fingered gripper, and a mobile base. As the focus of our experiments is on manipulating objects, we keep the base fixed for each episode.

We test the policy with client-server setup, where the policy is deployed as server, which on-robot client can query. The observations from the head camera of the robot are recorded on the client side, to maintain a recent history of 8 frames.  
Based on the average response time from the policy, we can execute predicted actions at approximately 5-7 Hz.

\section{\shortname for Long Horizon Tasks}

We show \shortname can perform long horizon tasks by using multiple prompt videos showing different tasks. To do so, we chain different prompt videos together to create a long horizon task. 
For example, \shortname can be used to prompt a robot to \textit{clean up the objects on a table into the drawer} by chaining prompt videos of 3 tasks: \textit{open the drawer}, \textit{place object on table into drawer}, and \textit{close the drawer}. 

\begin{figure*}
    \includegraphics[width=\linewidth]{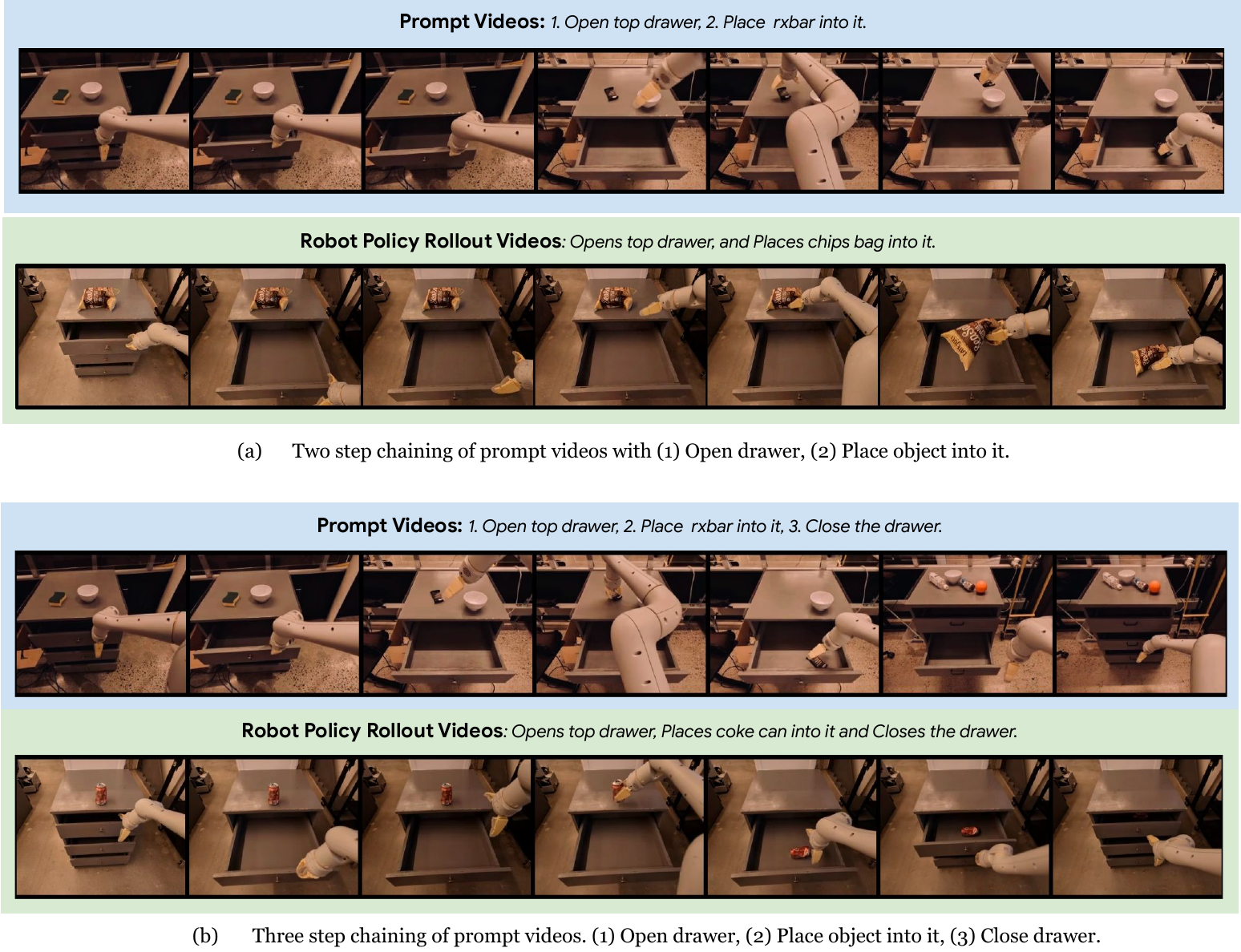}
    \caption{Qualitative results showing how \shortname can perform long horizon tasks with chaine prompt videos.}
    \label{fig:long_comp_policy_rollouts_combined}
\end{figure*}

When \shortname outputs terminate episode for the current prompt video, we change the prompt video to be the next sub-task without resetting the robot. In this manner, the robot can complete long horizon tasks with \shortname without explicitly being trained on long horizon videos. We show successful examples in Fig.~\ref{fig:long_comp_policy_rollouts_combined}. For videos of the robot completing long-horizon tasks, check the summary video in the supplementary material.

